\documentclass[journal]{IEEEtran}

\usepackage{amssymb}
\usepackage{amsmath,amsthm}
\usepackage{subfigure}
\usepackage{cite}
\usepackage{graphicx}
\usepackage{graphics}
\usepackage{color}
\usepackage{xspace}
\usepackage{psfrag}
\usepackage{algorithmicx}
\usepackage{algorithm}
\usepackage{algpseudocode}
\usepackage{multirow}
\usepackage{array}
\usepackage{url}
\usepackage[normalem]{ulem}
\usepackage{graphicx}
\usepackage{floatflt,setspace}
\usepackage{algcompatible}

\newtheorem{theorem}{Theorem}
\newtheorem{definition}{Definition}

\algnewcommand\INPUT{\item[\textbf{Input:}]}%
\algnewcommand\OUTPUT{\item[\textbf{Output:}]}%

\title{Unsupervised clustering under the Union of Polyhedral Cones (UOPC) model}
\author{Wenqi Wang, Vaneet Aggarwal, and Shuchin Aeron \thanks{W. Wang and V. Aggarwal are with the School of Industrial Engineering, Purdue University, West Lafayette, IN 47907. S. Aeron is with the Department of Electrical and Computer Engineering, Tufts University, Medford, MA  02155} }

\begin{document}
\maketitle
\begin{abstract}

In this paper, we consider clustering data that is assumed to come from one of finitely many pointed convex polyhedral cones. 
This model is referred to as the Union of Polyhedral Cones (UOPC) model. 
Similar to the Union of Subspaces (UOS) model where each data from each subspace is generated from a (unknown) basis, in the UOPC model each data from each cone is assumed to be generated from a finite number of (unknown) \emph{extreme rays}.
To cluster data under this model, we consider several algorithms - 
(a) Sparse Subspace Clustering by Non-negative constraints Lasso (NCL), 
(b) Least squares approximation (LSA), and 
(c) K-nearest neighbor (KNN) algorithm to arrive at affinity between data points.
Spectral Clustering (SC) is then applied on the resulting affinity matrix to cluster data into different polyhedral cones. 
We show that on an average KNN outperforms both NCL and LSA and for this algorithm we provide the deterministic conditions for correct clustering. For an affinity measure between the cones it is shown that as long as the cones are not very coherent and as long as the density of data within each cone exceeds a threshold, KNN leads to accurate clustering.
Finally, simulation results on real datasets (MNIST and YaleFace datasets) depict that the proposed algorithm works well on real data indicating the utility of the UOPC model and the proposed algorithm.


\end{abstract}
\vspace{-.1in}
\section{Introduction}

\begin{figure}[b]
\vspace{-.3in}
\includegraphics [ 
keepaspectratio, width=0.35\textwidth] {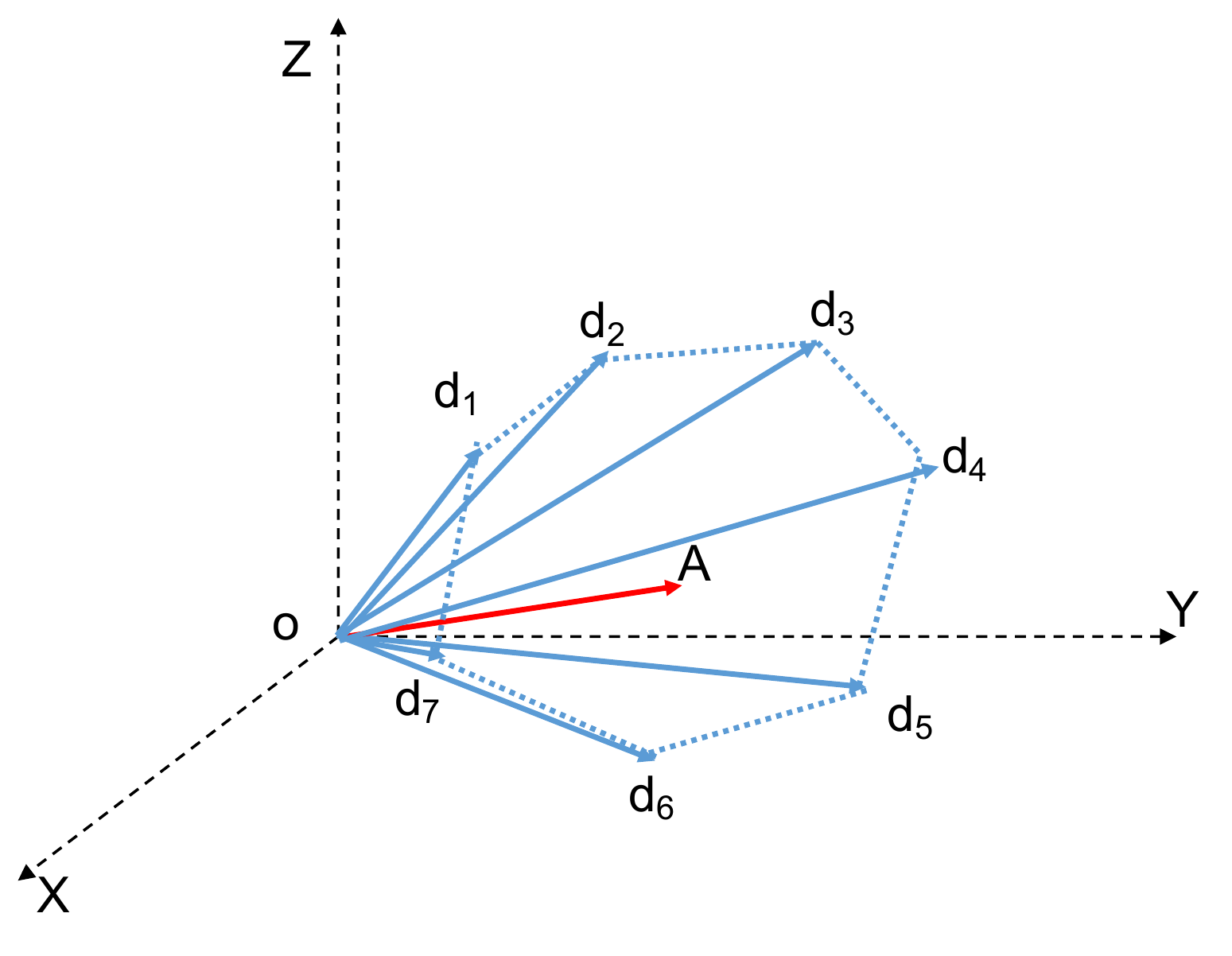}
\centering
\vspace{-.1in}
\caption{A polyhedral cone in 3D with 7 extreme rays: $od_1, od_2, od_3, od_4, od_5, od_6, od_7$. Any points inside the cone, e.g. $A$, can be represented by the non-negative linear combination of the extreme rays.} \label{eg_Cone}
\end{figure}

Clustering data which is assumed to come from the union of finitely many pointed convex polyhedral cones is motivated by several recent applications such as clustering human faces under different illuminance \cite{karmakar2015face,Li_2015_CVPR,ho2003clustering}, hyperspectral imaging \cite{dong2016hyperspectral,chen2013hyperspectral, li2014gabor}, metabolic network engineering \cite{oddsdottir2015dynamically,grigoriev2015algorithms}, and topic modeling \cite{min2010decomposing,kuang2015nonnegative}.
Clustering data under union of subspaces has been studied widely \cite{elhamifar2013sparse,soltanolkotabi2014robust,soltanolkotabi2012geometric}. The key difference with clustering data under the union of polyhedral cones is that given the extreme rays of a polyhedral cone\cite{wang2003illumination} as shown in Fig \ref{eg_Cone}, which serve as the generator of the polyhedral cone, the data inside the cone is a \emph{non-negative} linear combination of the extreme rays. 
{In other words a subspace is a cone (not pointed)  but a cone is not a subspace.}
Furthermore, an issue in extension of the approaches for sparse subspace clustering is that the extreme rays \emph{cannot} be represented as a positive linear combination of the other points in the same polyhedral cone, unlike the case of subspaces where there are no extreme rays. 
This in turn makes the \emph{identification} of the cone a harder problem in general and unless the data contains the extreme rays, it is computationally hard to uniquely identify the cone. This requirement is very similar to the case of non-negative matrix factorization \cite{hoyer2004non, lee2001algorithms}.

In this paper, we provide an method for clustering data which is assumed to come from the union of polyhedral cones. We first build an affinity graph with the different data points (treated as graph vertices) with the edge weights or affinities derived using a K-nearest neighbor (KNN) algorithm. Subsequently, spectral clustering \cite{ng2002spectral} is applied to obtain the correct clusters. We also note that algorithm along these lines have been considered in \cite{heckel2013robust,lucinska2012spectral}.  In \cite{heckel2013robust}, a thresholded subspace clustering (TSC) algorithm has been proposed for clustering data under the union of subspaces model, which uses nearest neighbors to create affinities that are then used to perform spectral clustering after thresholding.  The authors of \cite{lucinska2012spectral} employ mutual nearest neighbors are chosen to define affinity matrix. However, none of these works considered the problem of clustering when the data is assumed to come from a union of polyhedral cones (UOPC).  

Under the UOPC model, if for every extreme ray, the nearest point in other cone is farther than the $K^{\text{th}}$ nearest point in the same cone, then the graph will give $K$ true connections for each node and zero false connections. Thus, given enough \emph{density of data} in each polyhedral cone, the proposed approach will be able to give perfect clustering.  We formalize this by deriving deterministic conditions for clustering data in union of polyhedral cones. Clustering data under the UOPC model can also be performed by a modification of sparse subspace clustering (SSC) \cite{elhamifar2013sparse} with additional non-negative constraints for the coefficients. Such problems can be solved with Non-negative Constraint Lasso (NCL) algorithm \cite{itoh2015perfect}, which has a computational complexity of $O(N^3)$, where $N$ is the number of data points. As a special case of no sparsity condition, this algorithm has been considered in \cite{ho2003clustering}. However, theoretical analysis of these approaches for perfect clustering remain \emph{open}. With our proposed KNN-SC algorithm, we show that if there is enough density in the polyhedral cones and $K$ is large enough, the clustering will be perfect. Further,  applying KNN method to build the graph  takes $O(N\log N)$  using $kd\text{-}tree$ based KNN search\cite{moore1991intoductory}. Finally, through extensive sets of simulations, we show that the proposed approach performs better than the benchmark result proposed in \cite{ho2003clustering, itoh2015perfect} on both simulated and real datasets.

The rest of this paper is organized as follows. Section II describes the problem and Section III gives the proposed algorithm. Section IV gives the guarantees to the algorithm and numerical evaluations are presented in Section V. Finally, Section VI concludes this paper.

\vspace{-.05in}

\section{Problem Set-Up}
A pointed polyhedral cone ${\bf K}$ in $\mathbb{R}^n$ is defined as the set of all non-negative linear combination of the extreme rays $\bf{D}$, as one example shown in Fig \ref{eg_Cone},  thus
\begin{equation} \label{eq: Cones}
{\bf K} = \{{\bf D} {\bf a}: {\bf a}\in \mathbb{R}^{d \times 1} \geq 0, {\bf D} \in \mathbb{R}^{n\times d}\},
\end{equation}
where column vectors in ${\bf D}$ represent $d$ extreme rays that define the polyhedral cone ${\bf K}$. 

We are given a set of  $N$ data points, ${\bf X}_{i,i=1,2,...,N} \in \mathbb{R}^n$, uniformly sampled from a union of $L$ polyhedral cones, such that  ${\bf X}_i \in \bigcup _{\ell=1}^L {\bf K}^{(\ell)}$, where ${\bf K}^{(\ell)}$ is a pointed polyhedral cone with $d_\ell$ extreme rays in $\mathbb{R}^n$. The purpose is to find a partition to split $N$ data points into $L$ groups, denoted as $\zeta_1,\zeta_2,..., \zeta_L$,  such that data from each subgroups belong to the same polyhedral cones. \textbf{Note} that the number of extreme rays $d_\ell$ can be larger than the ambient dimensionality $n$.

\vspace{-.1in}
\section{Algorithm: KNN-SC}


The proposed algorithm, KNN-SC, is described in Algorithm \ref{mainalgo}. The data points are first normalized on a unit sphere. We consider two variants - denoted as the Gaussian Kernel Case, and the Binary Kernel Case. In these variants, the $K$ nearest neighbors are used for each data points and the affinity between them, summarized in ${\bf A}$, is based on the distance or $1$ in the two cases respectively. Finally, Spectral Clustering is performed on ${\bf A}^\top + {\bf A}$.

 
 \begin{algorithm}
   \caption{KNN-SC Algorithm}
  \begin{algorithmic}[1]
    \INPUT  Data points ${\bf X}_1, {\bf X}_2,..., {\bf X}_N$, number of clusters $L$, scaling parameter $\tau$, 
    maximum number of  neighbors $K$
    \OUTPUT Clustering Result $\zeta_1,\zeta_2,..., \zeta_L$
    \STATE Normalize each data by $\ell_2$ norm to get the normalized data  ${\bf Y}_i$ such that ${\bf Y}_i =\frac{{\bf X}_i}{\|{\bf X}_i\|_{\ell_2}},\forall_{i:i=1,2,...,N}$

	 	\FOR{$i \in \{1,2,...,N \}$}
		\STATE \text{\bf (Build affinity matrix $A$) }
		
		{\bf Gaussian Kernel Case: } ${\bf A}(i, j) = e^{-d^2({\bf Y}_i, {\bf Y}_j)/2\tau^2}$ if ${\bf Y}_j$ is  top-$K$ neighbor of ${\bf Y}_i$ and ${\bf A}_{i,j}=0$ otherwise.
		
		{\bf Binary Kernel Case: } ${\bf A}(i,j) = 1 $ if ${\bf Y}_j$ is  top-$K$ neighbor of ${\bf Y}_i$ and ${\bf A}_{i,j}=0$ otherwise.
		\ENDFOR
	\STATE ${\bf C} = {\bf  A}^\top + {\bf  A}$
	\STATE \text{\bf (Spectral Clustering) }Apply Spectral Clustering\cite{ng2002spectral} corresponding to ${\bf C}$ and get the partition $\zeta_1,\zeta_2,...., \zeta_L$
\end{algorithmic}
\label{mainalgo}
\end{algorithm}

\vspace{-.1in}
\section{Analysis Of KNN-SC}

We note that if we have enough points in each polyhedral cone, and the distance between polyhedral cones is large enough,  the algorithm will likely succeed in correctly clustering the data. In this section, we will formalize this notion. In order to do that, we first define the concept of false and true discoveries.


{\definition (False and True Discoveries \cite{soltanolkotabi2014robust}) } Fix $i$ and $j \in  \{1,2,3,..., i-1,i+1,...,N\}$, let ${\bf A}$ be the affinity matrix  built in step 3 of the KNN-SC Algorithm. Then $(i,j)$ obeying $A_{ij} \neq 0$ is  (i) a false discovery if $X_i$ and $X_j$ do not originate from the same polyhedral cone, and (ii) a true discovery if $X_i$ and $X_j$  originates from the same polyhedral cone.


When there are no false discoveries, points from different clusters will be well separated from each other as no connections exist between different clusters\cite{ng2002spectral}.
When there are many true discoveries in the affinity matrix, points from the same cluster will form a well connected graph and there will be no separations within data from the same clusters. 
Thus No False Discoveries and Many True Discoveries together indicate a perfect performance of the clustering algorithm. 

Applying KNN-SC algorithm under UOPC models, the pair-wise distance between any two polyhedral cones and the density inside each cone determine the accuracy of the clustering algorithm. Intuitively, the larger distance between the polyhedral cones and the denser the data from each cone,  the larger is the value of  $K$ which can be used to  satisfy no false discoveries and $K$ true discoveries for every data point thus obtaining many true connections. 

We next define the notion of affinity, which is a measure of distance between two polyhedral cones.
{\definition (Affinity)
Let ${\bf K}^{(i)}$ and ${\bf K}^{(j)}$ be the two polyhedral cones defined by \eqref{eq: Cones}, then the affinity between the two polyhedral cones is defined as
\begin{equation} \label{eq: Affinity2}
\text{Aff}({\bf K}^{(i)}, {\bf K}^{(j)}) =\min_{p,q} \|{\bf Y}_p^{(i)}-{\bf Y}_q^{(j)}\|_{\ell_2},
\end{equation}
where ${\bf Y}_p^{(i)}$ and ${\bf Y}_q^{(j)}$ are normalized (as unit $\ell_2$ norm) data from ${\bf K}^{(i)}$ and ${\bf K}^{(j)}$ respectively.}

The notion of affinity indicates the closest distance between the normalized data points in the two polyhedral cones thus defining a notion of distance between polyhedral cones. The affinity is bounded in the range [0,2]. We next define the Affinity Condition for the UOPC model, which indicates the minimum distance between any two polyhedral cones are separated by at least the threshold value $t^*$.
{\definition (Affinity Condition)
We say that the union of polyhedral cones obeys affinity condition with parameter $t^*$ if for any pair of polyhedral cones ${\bf K}^{(i)}$ and ${\bf K}^{(j)}$, $i\ne j$,
\begin{equation}\label{eq: AffinityCondition}
\text{Aff}({\bf K}^{(i)}, {\bf K}^{(j)}) \geq t^*.\end{equation}}

Next, we will introduce the concept of density of a polyhedral cone ${\bf K}^{(l)}$ at any location ${\bf v}$ drawn from the surface of a unit sphere in ${\mathbb R}^n$ with a paraeter $J$, denoted as  $\rho_l({\bf v},J)$. 
{\definition (Density)
\begin{equation} \label{eq: Density}
\rho_l(\mathbf{v},J) =\frac{J}{V_n(r_{{\bf v},J})},
\end{equation}
$r_{{\bf v},J}$ is the radius of $n$-sphere in $\mathbb{R}^n$ such that the ball centered at $\bf{v}$ with the radius $r_{{\bf v},J}$ includes exactly $J$ {\bf normalized} points from ${\bf K}^{(l)}$ (not including ${\bf v}$), 
$V_n(x) = \frac{\pi^{n/2}}{\Gamma (\frac{n}{2}+1)}{x^n}$ is the volume of a $n$-sphere. }

For data uniformly distributed in the cone, the density would be higher inside the cone as compared to that closer to the boundary. In order to have a single notion of desity for the cone, we define the density of the polyhedral cone ${\bf K}^{(l)}$  with parameter $J$ as

\begin{equation}
\rho_l(J) =\min _{\mathbf{v} \in {\bf K}^{(l)}} \rho_l(\mathbf{v},J).
\end{equation}


The condition that the cone is dense enough is denoted by the Density Condition given below.
{\definition (Density Condition)
We say that union of polyhedral cones obeys density condition with parameters $\rho^*(J)$ and $J$,  if the density $\rho_l(J)$ of any polyhedral cone ${\bf K}^{(l)}$ satisfies
\begin{equation}\label{eq: ana_sampling}
\rho_l(J) \geq \rho^*(J).
\end{equation}}
We have the following main result.
{\theorem (No False Discoveries) Under the union of L polyhedral cones from $\mathbb{R}^n$, assume that the affinity condition holds with parameter $t^*$ and the density condition holds with parameter $\rho^*(K)$. If $\rho^*(K) \geq \frac{K}{V_n(t^*)}$, then there will be no false discovery for the affinity matrix ${\bf A}$ that is constructed by KNN-SC with $K$ neighbors.}

\proof Let ${ P}$ be an arbitrary point selected from the cone ${\bf K}^{(l)}$. Based on affinity condition, within a ball centered at $P$ with radius of $t^*$, there are no points from the other cones. Since $\rho^*(K) V_n(t^*) \geq K$,  there are more than $K$ points from the cone ${\bf K}^{(l)}$ within the ball centered at $P$ with radius $t^*$. Thus the $K$ nearest neighbors are all from the cone ${\bf K}^{(l)}$ and there is No False Discovery.
\endproof

Theorem 1 gives a upper bound, $K \leq \rho^*(K) V_n(t^*)$, for finding the number of neighbors used in KNN algorithm. In order to satisfy many-true discoveries, we can have $K = \lfloor\rho^*(K) V_n(t^*)\rfloor$.

\vspace{-.1in}
\section{Numerical Results}


In this section, we will compare KNN-SC with other baselines for both synthetic and real datasets. The baseline we consider is a LASSO-based algorithm  \cite{itoh2015perfect}, which  solves the following optimization problem for each data point $y$, 
\begin{equation}
\min_{c_y} \frac{1}{2} \|{\bf y} - {\bf Y c_y}\|_2^2 + \lambda \mathbf{ 1}^\top {\bf c_y}  \quad \text{s.t.} \quad {\bf c_y}\geq {\bf 0},
\end{equation}
where ${\bf y}$ is any one of data, ${\bf Y}$ represents the remaining data not including ${\bf y}$, and ${\bf c_y}$ are the non-negative coefficients that represent ${\bf y}$ as a linear combination of ${\bf Y}$. Then, the coefficients  ${\bf c_y}$ for each data point ${\bf y}$ are combined in a matrix ${\bf C}$. Finally, Spectral Clustering is performed on ${\bf A}= {\bf C}+{\bf C}^\top$. We denote our Lasso-based clustering algorithm as NCL.  Further,  Least Square Approximation (LSA) proposed in \cite{ho2003clustering} to do clustering under UOPC is a special case of the above with $\lambda=0$. We note that the authors of \cite{ho2003clustering} compared different algorithms for YaleFace Dataset B \cite{georghiades2001few}, where it is found that the LSA algorithm  gives $6.67\%$ clustering error, which is the lowest among the compared algorithms. The compared algorithms  include K-means algorithm ($78.44\%$ error), Spectral Clustering (47.78\% error ), K-subspace (59.00\% error), and Spectral Curvature Clustering (58.00\% error). Thus, we use LSA for comparison, and do not include the other algorithms as baselines in this paper. 


\vspace{-.1in}
\subsection{Synthetic Data in 2-Dimension}
\begin{figure}[t!]
\includegraphics [trim=0.5in 2.6in 1in 2.8in, keepaspectratio, width=0.4\textwidth] {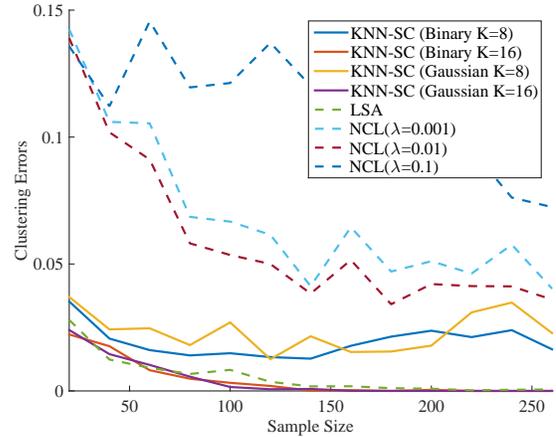}
\centering
\vspace{-.1in}
\caption{Clustering Error Versus Sample Size in each cone in 2-Dimensional Case.}
\label{simu_2D}
\vspace{-.2in}
\end{figure}
A cone in 2D is predefined by 2 fixed extreme rays. We  consider the data from two polyhedral cones, where the extreme rays corresponding to the first cone are $[\cos(-\frac{\pi}{2}),\sin (-\frac{\pi}{2})]$ and $[\cos \frac{2\pi}{9}, \sin \frac{2\pi}{9}]$,  and the extreme rays corresponding to the second cone are $[\cos \pi, \sin \pi]$ and $[\cos\frac{5\pi}{18}, \sin\frac{5\pi}{18}]$. 
Let there be $N$ data points from each polyhedral cone drawn randomly between the two extreme rays by a convex combination of the extreme rays where the coefficient is selected from a uniform distribution. 

Clustering error for different values of $N$ from $N=20$ to $N=260$ is shown in 
Figure \ref{simu_2D}. For each plotted point, the clustering error is the average result of 100 iterations. The numerical results show that KNN-SC algorithm with either Binary Kernels or Gaussian Kernels outperforms both the NCL and LSA algorithms.  LSA algorithm is comparable to KNN-SC algorithm, but it does not converge to zero clustering error. In contract, when the sample size is larger than $140$ in each polyhedral cone, KNN-SC algorithm with $K=16$ gives perfect clustering result, which is consistent with the No False Discoveries. 
For lower K, such as $K=8$, the clustering error is higher because of not enough true connections which leads to not enough connections holding a cluster together (which can be split into sub-cluster and a sub-cluster can get attached to a different cluster).

\vspace{-.1in}
\subsection{Synthetic Data in 3-Dimension}
\begin{figure}[ht]
\includegraphics [trim=0.5in 2.6in 1in 2.9in, keepaspectratio, width=0.4\textwidth] {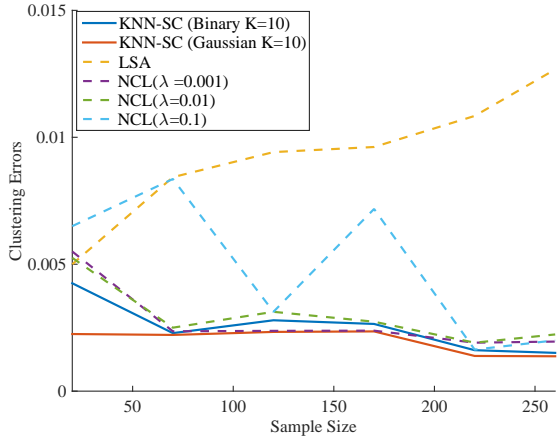}
\centering
\vspace{-.1in}
\caption{Clustering Error Versus Sample Size in each cone in 3-Dimensional Case}
\label{simu_3D}
\end{figure}
 Unlike in 2D scenario,  a polyhedral cone in 3D is defined by at least 3   extreme rays. To generate synthetic data, we only  pre-determined three extreme rays to build the cone. Represented by polar coordinates, the extreme rays corresponds to the first cone are $[1, \frac{2\pi}{9}, \frac{\pi}{4} ]$, $[1, \frac{\pi}{2}, \frac{\pi}{4} ]$, $[1, 0, 0 ]$, while the the extreme rays corresponds to the second cone are $[1, \frac{5\pi}{18}, \frac{\pi}{4} ]$, $[1, \frac{\pi}{2}, \frac{\pi}{2} ]$, $[1, \frac{\pi}{4}, \frac{\pi}{2} ]$. Let there be $N$ data points from each cone and data inside the the cone is generated by the non-negative affine combination of the three corresponding extreme rays, where the three coefficients are generated by standard entry-wise uniform distribution between 0 and 1 which are then  normalized by the sum the three coefficients.

Clustering error for different values of $N$ from $20$ to $270$ is shown in Figure 3. For each plotted point, the clustering error is the average result of 100 iterations. The numerical results show that when sample size is large than $70$, the NCL algorithm with proper parameters, such as $\lambda =0.01$ or $\lambda  = 0.001$  performs slightly worse as compared to the KNN-SC algorithm with Gaussian Kernels. LSA performs the worst in this case, unlike its good performance in synthetic data in 2-dimensional scenario. For KNN-SC algorithm, Gaussian Kernel performs better than the Binary Kernel for the parameters selected.

\vspace{-.1in}
\subsection{MNIST Dataset}
\begin{figure}[t!]
\includegraphics [trim=0.5in 2.6in 1in 2.8in, keepaspectratio, width=0.4\textwidth] {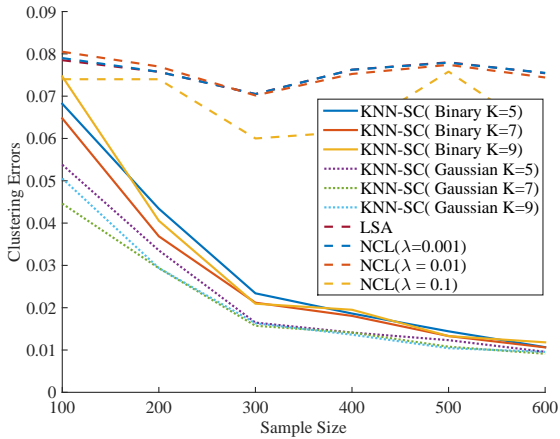}
\centering
\vspace{-.1in}
\caption{Clustering Error Versus Sample Size for each digit in MNIST dataset}
\label{simu_MNIST}
\vspace{-.1in}
\end{figure}

We consider two digits, $1$ and $2$ in the handwritten digits from MNIST dataset \cite{mnistlecun}, and compare the performance of the algorithms for varying amount of data points $N$ in each cluster from $N=100$ to $N=600$ sampled uniformly. The comparison is depicted in 
Figure \ref{simu_MNIST}, where each data point is the average over 100 random selections of points. We note that  KNN-SC algorithm with Gaussian kernels performs better than KNN-SC with Binary kernels. 
Further,  clustering error decreases when $K$ increases from $5$ to $7$, which is caused the increasing number of true discoveries. Further clustering error decreases when  $K$ increases from $7$ to $9$ since the number of false discoveries increase.  KNN-SC algorithm with $K=7$ and Gaussian kernels performs better than the other algorithms, with  clustering error less than  1\% when the sample size is $600$ for each digit.


\vspace{-.1in}
\subsection{YaleFace Dataset}
\begin{figure}[t!]
\includegraphics [trim=0.5in 2.6in 1in 2.8in, keepaspectratio, width=0.4\textwidth] {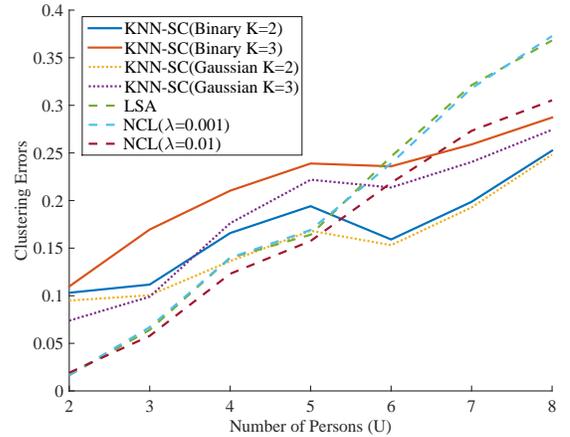}
\centering
\vspace{-.1in}
\caption{Clustering Error Versus Number of Persons in YaleFace Dataset.}
\label{simu_YALE}
\vspace{-.1in}
\end{figure}


We will now evaluate the performance of KNN-SC algorithm in clustering human faces from the Extended Yale B data set \cite{georghiades2001few, lee2005acquiring}. We resize the images to  $48 \times 42$ pixels for ease of calculation. YaleFace includes the faces of $38$ persons with $64$ faces for each person. 
We choose 
$U$ out of $38$ people uniformly at random and depict the comparisons of different algorithms for varying values of $U$ from two to eight in 
Figure \ref{simu_YALE}.  For each point plotted, the clustering error is the average of $20$ random choices of selecting $U$ persons. We note that NCL and LSA algorithms are comparable, and perform the best when number of persons selected is small. The clustering error increases as the number of persons ($U$) increases, and KNN-SC performs the best for $U>5$. Beyond $U=6$, there is more than $10\%$ gap between the performance of KNN-SC with $K=2$ and Gaussian Kernels and the LSA/NCL algorithms.


\vspace{-.1in}
\section{Conclusion}
This paper proposes an algorithm, KNN-SC, to cluster data into union of polyhedral cones by applying spectral clustering algorithm on $K$-Nearest-Neighbour based graph.  Deterministic guarantees on the algorithm performance are provided. Furthermore, our analysis can be extended to  polytope clustering model. Extensive set of experiments depict that the proposed algorithm works well in both synthetic and realistic settings. 
\bibliographystyle{IEEEtran}
\bibliography{Ref}
\end{document}